\documentclass[lettersize,journal]{IEEEtran}
\usepackage{amsmath,amsfonts,amssymb}
\usepackage{array}
\usepackage{subcaption}
\usepackage{textcomp}
\usepackage{stfloats}
\usepackage{url}
\usepackage{xcolor}
\usepackage{hyperref}
\usepackage{verbatim}
\usepackage{graphicx}
\usepackage{algorithm}
\usepackage{algpseudocode}
\algrenewcommand\algorithmicrequire{\textbf{Input:}}
\algrenewcommand\algorithmicensure{\textbf{Output:}}
\usepackage{bm}
\def\BibTeX{{\rm B\kern-.05em{\sc i\kern-.025em b}\kern-.08em
    T\kern-.1667em\lower.7ex\hbox{E}\kern-.125emX}}
\usepackage{balance}
\begin{document}
\title{Patch-Based Deep Kernel Learning for Data Assimilation and Bias Correction with Extremely Sparse Observations}

\author{Emanuele  Mele,  \IEEEmembership{Student  Member,  IEEE,}
        Angelo  Coluccia,  \IEEEmembership{Senior  Member,  IEEE,}    
        Massimo  Cafaro,   \IEEEmembership{Senior  Member,  IEEE,}
        Italo  Epicoco
        and  Paola Nassisi
\thanks{E. Mele, A. Coluccia, M. Cafaro and I. Epicoco are with the Department of Engineering for Innovation, University of Salento, 73100 Lecce, Italy(e-mail: name.surname@unisalento.it); E. Mele, M. Cafaro, I. Epicoco and P. Nassisi are with the Euro-Mediterranean Center on Climate Change, Foundation (CMCC), 73100 Lecce, Italy (e-mail: name.surname@cmcc.it).}
}

\markboth{IEEE Transactions on Big Data,  Vol.  XX, No.  XX, Month  2026}%
{Mele \MakeLowercase{\textit{et al.}}: Patch-Based Deep Kernel Learning for Data Assimilation and Bias Correction with Extremely Sparse Observations}

\maketitle

\begin{abstract}
Data Assimilation (DA) reconstructs the state of a spatio-temporal system by combining sensor observations with the output of a numerical model. Since model states are high-dimensional while observations cover only a small fraction of the domain, DA is a challenging large-scale estimation problem, central to weather forecasting, oceanography and many other partially observed systems. Classical variational methods such as 3D-Var rely on a hand-crafted background error covariance that assumes unbiased errors, and therefore cannot capture state-dependent correlations or systematic model biases. Moreover, they compute the analysis over the full state space, so their cost grows with the size of the domain. Deep learning (DL) methods can perform DA and bias correction (BC) non-linearly, but standard architectures typically offer limited interpretability and do not natively provide calibrated uncertainty estimates. We propose a hybrid framework that couples a convolutional neural network (CNN) with a learnable-kernel Gaussian Process (GP) to jointly perform DA and BC from extremely sparse observations. The model is trained end-to-end by maximizing the marginal log-likelihood, learns a data-driven covariance over the whole grid, and provides both the Maximum A Posteriori analysis and its uncertainty by design. A patch-based strategy reduces the $\mathcal{O}(N^3)$ training cost of exact GPs over $N$ grid points to $\mathcal{O}(N)$ for a fixed patch size. Inference remains exact and only requires solving a linear system whose $\mathcal{O}(m^3)$ cost depends on the number of observations $m \ll N$, not on the domain. The learned covariance can also replace the hand-crafted one in a classical 3D-Var scheme, improving its accuracy without changing the variational formulation. Experiments on sea surface temperature and salinity, with both synthetic and real, extremely sparse in-situ observations, show that the framework outperforms 3D-Var, a classical GP and a competing DL model, while reducing inference time by more than 98\% with respect to 3D-Var.
\end{abstract}


\section{Introduction}

Data Assimilation (DA) estimates the state of a dynamical system by combining observations with the prior information provided by a numerical model \cite{DAFound,bannister2017review,cheng2023mldauq}. More generally, it is a state-estimation and data-fusion problem in which heterogeneous, noisy and sparse measurements are integrated with a model-based representation of the system \cite{zheng2015fusion,khan2008kalman,sun2018kalmantraffic}. A central component is bias correction: numerical models suffer from systematic errors due to unresolved processes, discretization and parametrization choices, which must be identified and removed \cite{biascorr,biascorr2,farchi2021comparison}. Beyond numerical weather prediction and geosciences, related formulations arise in land-surface modelling, air-quality estimation, traffic monitoring and other partially observed high-dimensional systems \cite{cheng2023mldauq,li2024landda,raoult2025daML,zhu2017airquality,le2017localgp}, where combining high-volume, heterogeneous observation streams with a model prior is a defining challenge of the big-data era \cite{chen2014bigdata,liang2024stbigdata}.

The problem is particularly hard when observations are sparse relative to the state dimension. Sensors cover only a small subset of locations, while the state and its bias must be reconstructed over the entire domain \cite{manohar2018sensor,joneidi2020sensor,liu2018lsdecomp}. How information propagates from observed to unobserved regions is largely determined by the error covariance structure, whose specification is central to both variational and ensemble-based DA \cite{bannister2017review,lorenc2003covariance} and remains difficult for high-dimensional data in general \cite{wen2021covariance,chu2018pmu}.

Variational methods such as 3D-Var and 4D-Var balance the deviation from a background state against the discrepancy with the observations, and their effectiveness depends critically on the background-error covariance matrix \cite{bannister2017review,lorenc2003covariance}. Standard formulations also assume unbiased errors, which limits their ability to represent systematic, spatially coherent biases. Ensemble Kalman filters represent uncertainty through an ensemble of states, but the relatively small ensemble sizes affordable in large systems lead to sampling errors in the estimated covariances, and the computational cost remains high \cite{lorenc2003enkf,bannister2017review,khan2008kalman}. The cost of estimating and updating high-dimensional uncertainty and bias structures mirrors the scalability bottlenecks of probabilistic models on big data \cite{chen2014bigdata,liu2020scalablegps}.

Machine Learning (ML) offers a complementary direction \cite{convml,reichstein2019deep,zhu2017dlremote}. Neural networks learn nonlinear representations directly from data and have been used for spatio-temporal mining and forecasting \cite{wang2022stdm,ji2025weather,gong2024weather,zhang2020rainfall}, data imputation \cite{yuan2023stgan,shi2018missing} and ocean state reconstruction \cite{meng2023physicsgan,meng2022subsurface,fablet2023multimodal}. Within DA, deep learning has been applied to latent-space assimilation, observation operators, model-error and bias correction, and end-to-end variational schemes \cite{arcucci2021deep,amendola2021data,shanh2024dnnobs,beauchamp2022fourdvarnet,febvre2022joint,farchi2023online,xu2025fuxida,lean2026,diffda,geer2021learning}, while Bayesian deep learning has shown the value of calibrated uncertainty in environmental forecasting \cite{han2022bayesiandl}.

Purely neural approaches, however, provide accurate corrections, including implicit bias removal, but little information about their uncertainty or about how each observation influences the reconstructed state and bias. In sparse regimes, where most of the field is inferred rather than observed, uncertainty quantification is essential to assess the reliability of the corrections and to distinguish well-supported updates from extrapolations \cite{cheng2023mldauq,raoult2025daML,hullermeier2021uncertainty,mallick2024uq}.

Gaussian Processes (GPs) naturally address part of this challenge, as they provide predictive means and covariances as direct outputs of inference \cite{gpfmlmit}. However, their expressiveness is bounded by predefined kernels, which struggle to capture state-dependent relationships and spatially varying biases, and exact inference scales as $\mathcal{O}(n^3)$, a limitation that becomes severe for large sensor networks and high-resolution grids \cite{hensman2013bigdata,liu2020scalablegps,titsias2009variational,gardner2018gpytorch}. Local and patch-based approximations mitigate this cost \cite{le2017localgp,park2016patching,park2018patchwork,gu2012spatialgp}, while deep kernel learning lets the kernel operate on neural features, increasing its flexibility while retaining principled uncertainty quantification \cite{wilson2016deep,wilson2015kissgp}. For DA, this yields a data-adaptive covariance able to jointly represent bias correction and error propagation.

In this work, we exploit this hybrid perspective for extremely sparse DA, jointly treating state and bias correction. We couple a convolutional neural network with a learnable-kernel GP, trained end-to-end by maximizing the marginal log-likelihood. The network learns a representation of the background state that captures systematic bias patterns, and the GP uses it to build a data-dependent background-error covariance. Unlike classical variational approaches, spatial dependencies and bias structures are learned rather than imposed; unlike deterministic deep learning, predictive uncertainty is an intrinsic output. The analysis is obtained via Maximum A Posteriori (MAP) estimation, which here coincides with the Minimum Mean Square Error (MMSE) estimate. The model learns how a few observations propagate to unobserved locations, and a spatial patching strategy limits the covariance computations during training while preserving global reconstruction at inference.

We use oceanographic data as case study, since ocean fields exhibit strong spatial dependencies, nonlinear interactions, heterogeneous observations and well-documented systematic biases, while in-situ measurements cover only a small fraction of the domain \cite{meng2022subsurface, fablet2023multimodal}. The framework we propose is, however, fully general, as it can be adopted to the estimation of any high-dimensional field and its bias from model state and few observations, a problem formulation shared by many other applications \cite{cheng2023mldauq,li2024landda,raoult2025daML,zhu2017airquality}.

The contributions of this work are as follows. First, we introduce a hybrid CNN/GP architecture for DA in which the covariance, and hence the bias correction, is learned from the latent representation of the background state. Second, we provide a probabilistic formulation that yields both a MAP correction, including bias removal, and spatially predictive uncertainty. Third, we propose a patch-based strategy that reduces the computational and memory cost of exact GP training, making the framework applicable to large, high-resolution domains with a controllable trade-off between cost and accuracy. Fourth, we show that the learned covariance can be used directly as the background-error covariance of a classical 3D-Var scheme, consistently improving over the hand-crafted Gaussian covariance without additional tuning. Finally, we evaluate the framework on Sea Surface Temperature (SST) and Sea Surface Salinity (SSS) in a domain that includes the Adriatic, Tyrrhenian and Ionian Seas, using both synthetic observations with controlled density and real, irregularly distributed in-situ measurements.


\section{Background}
\label{background}

\begin{figure*}
    \centering
    \includegraphics[width=\linewidth]{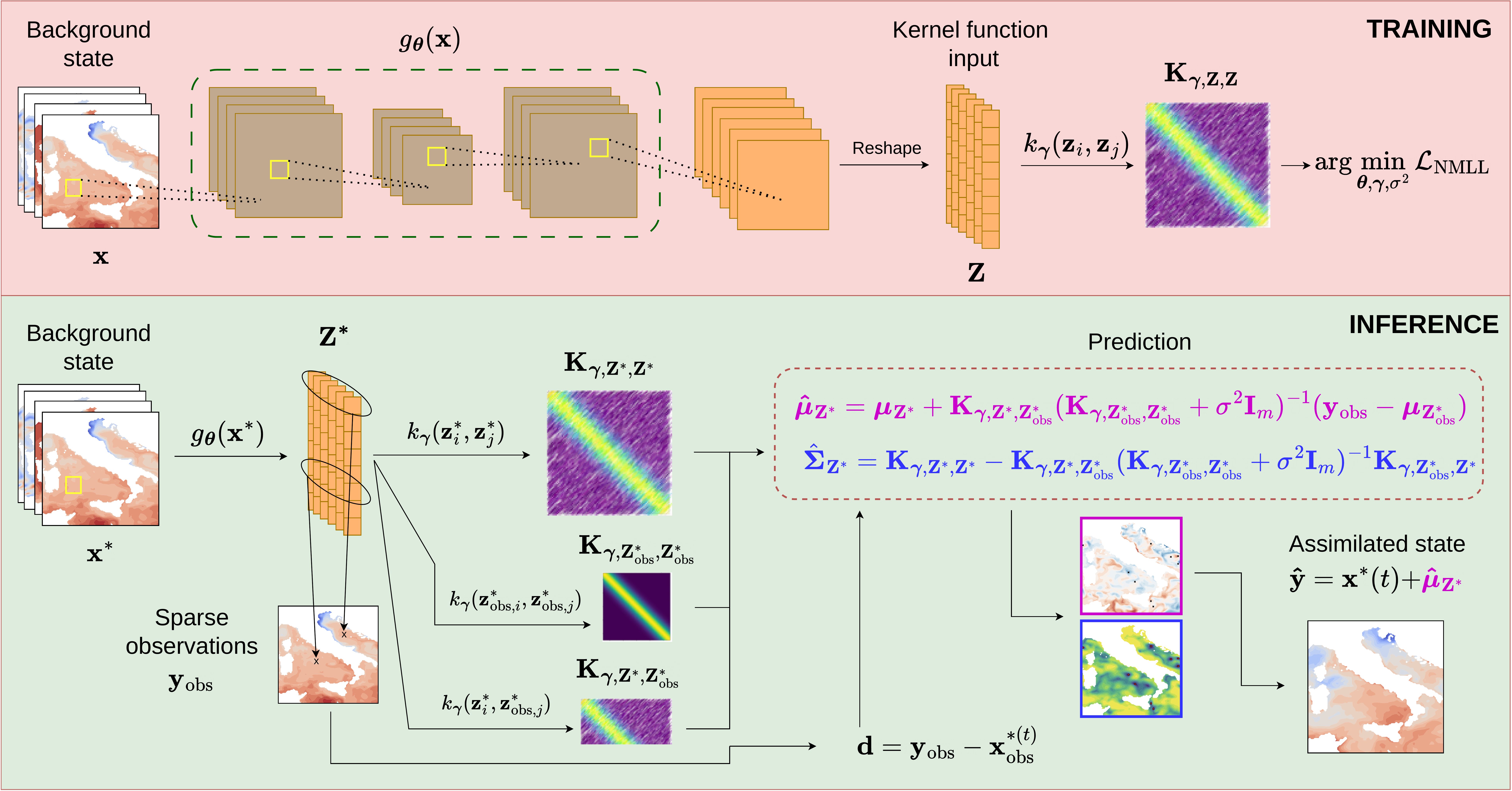}
    \caption{Proposed data assimilation framework combining a neural network and a Gaussian process. Top: During training, the neural network learns an embedding $\mathbf{Z}$ of the input $\mathbf{x}$, which is fed to the learnable kernel $k_{\boldsymbol{\gamma}}$ to compute the covariance matrix over the grid and optimize the marginal log-likelihood. Bottom: At inference, the embeddings $\mathbf{Z}^*$ of the background state are used to compute the covariance matrices between observed and unobserved locations; conditioned on the residuals at the observed positions, the model predicts the background error corrections $\hat{\boldsymbol{\mu}}_{\mathbf{Z}^*}$ and their uncertainties $\hat{\boldsymbol{\Sigma}}_{\mathbf{Z}^*}$, which are added to the background state to obtain the assimilated field.}
    \label{fig:arch}
\end{figure*}

We briefly review GPs following the notation of \cite{gpfmlmit}. Let $\mathbf{X} = [\mathbf{x}_1, \dots, \mathbf{x}_n] \in \mathbb{R}^{D \times n}$ collect the $n$ training inputs and $\mathbf{y} \in \mathbb{R}^{n}$ the corresponding scalar targets. Placing a GP prior on a latent function $f$, the vector of function values $\mathbf{f} = [f(\mathbf{x}_1), \dots, f(\mathbf{x}_n)]^\top$ is jointly Gaussian, ${\mathbf{f} \sim \mathcal{N}(\boldsymbol{\mu}_{\mathbf{X}}, \mathbf{K}_{\boldsymbol{\gamma},\mathbf{X},\mathbf{X}})}$, where $\boldsymbol{\mu}_{\mathbf{X}} \in \mathbb{R}^{n}$ is the prior mean vector and $\mathbf{K}_{\boldsymbol{\gamma},\mathbf{X},\mathbf{X}} \in \mathbb{R}^{n \times n}$ is the covariance matrix with entries $[\mathbf{K}_{\boldsymbol{\gamma},\mathbf{X},\mathbf{X}}]_{i,j} = k_{\boldsymbol{\gamma}}(\mathbf{x}_i, \mathbf{x}_j)$, $k_{\boldsymbol{\gamma}}$ being a kernel with hyperparameters $\boldsymbol{\gamma}$. For $n^*$ test inputs $\mathbf{X}^* = [\mathbf{x}_1^*, \dots, \mathbf{x}_{n^*}^*] \in \mathbb{R}^{D \times n^*}$, the prior mean $\boldsymbol{\mu}_{\mathbf{X}^*} \in \mathbb{R}^{n^*}$, the covariance $\mathbf{K}_{\boldsymbol{\gamma},\mathbf{X}^*,\mathbf{X}^*} \in \mathbb{R}^{n^* \times n^*}$ and the cross-covariance $\mathbf{K}_{\boldsymbol{\gamma},\mathbf{X}^*,\mathbf{X}} \in \mathbb{R}^{n^* \times n}$ are defined analogously, the subscripts indicating the inputs involved.

Assuming additive Gaussian noise, the observation model is $\mathbf{y} = \mathbf{f} + \boldsymbol{\epsilon}$ with $\boldsymbol{\epsilon} \sim \mathcal{N}(\mathbf{0}, \sigma^2 \mathbf{I}_n)$. Denoting by ${\mathbf{K}_{\mathbf{y}} = \mathbf{K}_{\boldsymbol{\gamma},\mathbf{X},\mathbf{X}} + \sigma^2 \mathbf{I}_n}$ the covariance of the targets and by $\mathbf{r} = \mathbf{y} - \boldsymbol{\mu}_{\mathbf{X}}$ the residual with respect to the prior mean, the predictive distribution of $\mathbf{f}^* = [f(\mathbf{x}_1^*), \dots, f(\mathbf{x}_{n^*}^*)]^\top$ is $\mathbf{f}^* \mid \mathbf{X}^*, \mathbf{X}, \mathbf{y}, \boldsymbol{\gamma}, \sigma^2 \sim \mathcal{N}(\hat{\boldsymbol{\mu}}^*, \hat{\boldsymbol{\Sigma}}^*)$, with
\begin{align}
\hat{\boldsymbol{\mu}}^* &= \boldsymbol{\mu}_{\mathbf{X}^*} + \mathbf{K}_{\boldsymbol{\gamma},\mathbf{X}^*,\mathbf{X}} \mathbf{K}_{\mathbf{y}}^{-1} \mathbf{r}, \\
\hat{\boldsymbol{\Sigma}}^* &= \mathbf{K}_{\boldsymbol{\gamma},\mathbf{X}^*,\mathbf{X}^*} - \mathbf{K}_{\boldsymbol{\gamma},\mathbf{X}^*,\mathbf{X}} \mathbf{K}_{\mathbf{y}}^{-1} \mathbf{K}_{\boldsymbol{\gamma},\mathbf{X},\mathbf{X}^*}.
\end{align}
Since the Gaussian posterior is symmetric and unimodal, $\hat{\boldsymbol{\mu}}^*$ is both the MAP and the Minimum Mean Squared Error (MMSE) estimate.

The hyperparameters $\boldsymbol{\gamma}$ and $\sigma^2$ are estimated by maximizing the marginal likelihood $p(\mathbf{y} \mid \mathbf{X}, \boldsymbol{\gamma}, \sigma^2) = \mathcal{N}(\mathbf{y} \mid \boldsymbol{\mu}_{\mathbf{X}}, \mathbf{K}_{\mathbf{y}})$, whose logarithm is
\begin{equation}
\log p(\mathbf{y} \mid \mathbf{X}, \boldsymbol{\gamma}, \sigma^2) = -\frac{1}{2} \mathbf{r}^\top \mathbf{K}_{\mathbf{y}}^{-1} \mathbf{r} - \frac{1}{2} \log |\mathbf{K}_{\mathbf{y}}| + \text{const}.
\end{equation}
Both the linear system $\mathbf{K}_{\mathbf{y}}^{-1} \mathbf{r}$ and the log-determinant $\log |\mathbf{K}_{\mathbf{y}}|$ are typically computed via the Cholesky decomposition of $\mathbf{K}_{\mathbf{y}}$, which requires $\mathcal{O}(n^3)$ operations and $\mathcal{O}(n^2)$ memory. Once this is done, the predictive mean and variance cost $\mathcal{O}(n)$ and $\mathcal{O}(n^2)$ for each point, respectively.

\section{Proposed Method}
\label{sec:methodology}

The proposed methodology performs data assimilation by integrating deep learning with GP modeling. As depicted in Figure \ref{fig:arch}, once the model is trained, the framework consists of two stages: feature extraction from the composite background state via a CNN, followed by GP-based inference to estimate and correct the background error field.

Let the background state be multi-variable data generated by a prior model over a regular grid of $H \times W$ points, where $H$ and $W$ are the numbers of grid points along latitude and longitude, respectively, and $C$ is the number of channels (i.e., fields). Each field is vectorized so that its $i$-th entry, $i = 1, \dots, HW$, corresponds to the $i$-th grid point, and all fields are stacked into the vector $\mathbf{x} \in \mathbb{R}^{HWC}$. We partition it as $\mathbf{x} = [\mathbf{x}^{(t)\top}, \mathbf{x}^{(a)\top}]^\top$, where $\mathbf{x}^{(t)} \in \mathbb{R}^{HW}$ is the variable to be assimilated (e.g., temperature) and $\mathbf{x}^{(a)} \in \mathbb{R}^{HW(C-1)}$ collects the auxiliary fields (e.g., other oceanographic variables such as salinity and velocity components), which provide contextual information to the feature extractor. Given the ground-truth field $\mathbf{y}_{\text{gt}} \in \mathbb{R}^{HW}$ of the target variable, the quantity reconstructed during training is the background error, or residual field, $\mathbf{y} = \mathbf{y}_{\text{gt}} - \mathbf{x}^{(t)} \in \mathbb{R}^{HW}$, defined over the entire grid.

Since this target covers the whole domain rather than just the observation locations, the framework learns to reconstruct the full background error, which combines a spatially coherent systematic bias, persisting over large regions regardless of observability, and localized, higher-frequency deviations tied to individual measurements. The learned covariance does not separate these two components but captures both in a single data-driven correlation structure. With extremely sparse observations, the GP predictive mean is dominated by the learned prior mean, which encodes a global, low-frequency bias pattern, performing an implicit bias correction even far from any observation; as the density increases, the same mechanism adds localized corrections driven by the nearby measurements. This unified formulation allows the framework to jointly perform DA and bias correction without explicitly disentangling the two error components during training.

\subsection{Model setup and Training stage}
Rather than a standard isotropic covariance over the raw input space, the GP kernel operates on the features extracted by a neural network $g_{\boldsymbol{\theta}}(\cdot)$ with learnable parameters $\boldsymbol{\theta}$. The feature extractor is a CNN that processes the background state in its gridded form, i.e., as an $H \times W$ image with $C$ channels, fusing the target variable with the auxiliary fields. It outputs an $H \times W$ feature map with $h$ channels, i.e., an $h$-dimensional embedding $\mathbf{z}_i \in \mathbb{R}^h$ for each grid point $i = 1, \dots, HW$, which we collect column-wise in the matrix
\begin{align}
\mathbf{Z} = g_{\boldsymbol{\theta}}(\mathbf{x}) = [\mathbf{z}_1, \dots, \mathbf{z}_{HW}] \in \mathbb{R}^{h \times HW},
\end{align}
where, with a slight abuse of notation, $g_{\boldsymbol{\theta}}(\mathbf{x})$ includes the reshaping of $\mathbf{x}$ onto the grid and of the output feature map into $\mathbf{Z}$. The covariance between grid points is obtained by applying a base kernel $k_{\boldsymbol{\gamma}}$, e.g., Radial Basis Function (RBF), in the learned feature space:
\begin{align}
[\mathbf{K}_{\boldsymbol{\gamma},\mathbf{Z},\mathbf{Z}}]_{i,j} = k_{\boldsymbol{\gamma}}(\mathbf{z}_i, \mathbf{z}_j), \quad i,j = 1, \dots, HW.
\end{align}
The prior over the full background error field is then $$\mathbf{y} \mid \mathbf{x}, \boldsymbol{\theta}, \boldsymbol{\gamma}, \sigma^2 \sim \mathcal{N}(\boldsymbol{\mu}_{\mathbf{Z}}, \mathbf{K}_{\boldsymbol{\gamma},\mathbf{Z},\mathbf{Z}} + \sigma^2 \mathbf{I}_{HW})$$ where $\sigma^2$ is the observation noise variance and $\boldsymbol{\mu}_{\mathbf{Z}} \in \mathbb{R}^{HW}$ is the prior mean vector computed from the embeddings. The hyperparameters $\boldsymbol{\gamma}$ include the parameters of both the kernel and the prior mean (see Section~\ref{subsec:nn_gp_setup}). Note that both $\mathbf{K}_{\boldsymbol{\gamma},\mathbf{Z},\mathbf{Z}}$ and $\boldsymbol{\mu}_{\mathbf{Z}}$ depend on $\boldsymbol{\theta}$ through $\mathbf{Z}$.

During offline training, the residual field $\mathbf{y}$ is known over the entire grid, and the CNN parameters $\boldsymbol{\theta}$ and GP hyperparameters $(\boldsymbol{\gamma}, \sigma^2)$ are optimized jointly end-to-end by minimizing the Negative Marginal Log-Likelihood (NMLL):
\begin{align}
\mathcal{L}_{\text{NMLL}}(\boldsymbol{\theta}, \boldsymbol{\gamma}, \sigma^2) = \tfrac{1}{2}\, \mathbf{r}^\top \mathbf{K}_{\mathbf{y}}^{-1} \mathbf{r} + \tfrac{1}{2} \log |\mathbf{K}_{\mathbf{y}}|,
\end{align}
where $\mathbf{r} = \mathbf{y} - \boldsymbol{\mu}_{\mathbf{Z}}$ and $\mathbf{K}_{\mathbf{y}} = \mathbf{K}_{\boldsymbol{\gamma},\mathbf{Z},\mathbf{Z}} + \sigma^2 \mathbf{I}_{HW}$. Gradients are backpropagated through $\mathbf{K}_{\mathbf{y}}$ and $\boldsymbol{\mu}_{\mathbf{Z}}$ to update both $\boldsymbol{\theta}$ and $(\boldsymbol{\gamma}, \sigma^2)$. The training is summarized in Algorithm \ref{alg:training}.

\begin{algorithm}
\caption{Training of the proposed framework}
\label{alg:training}
\begin{algorithmic}[1]
\Require Background states $\mathbf{x}$ and ground truths $\mathbf{y}_{\text{gt}}$ of the training set, patch size $P$, maximum number of patches per step $b$, epochs $E$
\Ensure Trained parameters $\boldsymbol{\theta}$, $\boldsymbol{\gamma}$, $\sigma^2$
\State Initialize $\boldsymbol{\theta}$, $\boldsymbol{\gamma}$, $\sigma^2$
\For{$e = 1, \dots, E$}
    \For{each mini-batch of pairs $(\mathbf{x}, \mathbf{y}_{\text{gt}})$}
        \State $\mathbf{Z} \gets g_{\boldsymbol{\theta}}(\mathbf{x})$ \Comment{$\mathbf{Z} \in \mathbb{R}^{h \times HW}$}
        \State $\mathbf{y} \gets \mathbf{y}_{\text{gt}} - \mathbf{x}^{(t)}$ \Comment{background error}
        \State Partition the grid into $K$ patches of $P \times P$
        \State Sample at most $b$ patches
        \For{each sampled patch $k$}
            \State $\mathbf{Z}_k, \mathbf{y}_k \gets$ patch $k$ of $\mathbf{Z}$, $\mathbf{y}$
            \State $\mathbf{K}_{\boldsymbol{\gamma},\mathbf{Z}_k,\mathbf{Z}_k} \gets [k_{\boldsymbol{\gamma}}(\mathbf{z}_i, \mathbf{z}_j)]_{i,j}$
            \State $\boldsymbol{\mu}_{\mathbf{Z}_k} \gets$ prior mean at $\mathbf{Z}_k$
            \State $\mathbf{K}_{\mathbf{y}_k} \gets \mathbf{K}_{\boldsymbol{\gamma},\mathbf{Z}_k,\mathbf{Z}_k} + \sigma^2 \mathbf{I}_{P^2}$
            \State $\mathbf{r}_k \gets \mathbf{y}_k - \boldsymbol{\mu}_{\mathbf{Z}_k}$
            \State $\mathcal{L}^{(k)}_{\text{NMLL}} \gets \tfrac{1}{2}\mathbf{r}_k^\top \mathbf{K}_{\mathbf{y}_k}^{-1}\mathbf{r}_k + \tfrac{1}{2}\log|\mathbf{K}_{\mathbf{y}_k}|$
        \EndFor
        \State Update $\boldsymbol{\theta}, \boldsymbol{\gamma}, \sigma^2$ via $\nabla \sum_k \mathcal{L}^{(k)}_{\text{NMLL}}$
    \EndFor
\EndFor
\end{algorithmic}
\end{algorithm}

\subsection{Inference stage and Data Assimilation}
\label{inf_and_da}
Once trained, $\boldsymbol{\theta}$, $\boldsymbol{\gamma}$ and $\sigma^2$ are frozen. At inference, the ground truth is unavailable and the embeddings are computed from the current background state $\mathbf{x}^*$, i.e., ${\mathbf{Z}^* = g_{\boldsymbol{\theta}}(\mathbf{x}^*) \in \mathbb{R}^{h \times HW}}$. Observations $\mathbf{y}_{\text{obs}} \in \mathbb{R}^m$ are provided only at $m \ll HW$ grid points; we denote by ${\mathbf{Z}^*_{\text{obs}} \in \mathbb{R}^{h \times m}}$ the columns of $\mathbf{Z}^*$ at these points and by $\mathbf{x}^{*(t)}_{\text{obs}} \in \mathbb{R}^m$ the corresponding entries of the target channel $\mathbf{x}^{*(t)}$. The background error at the observed points is estimated by the departure $\mathbf{d} = \mathbf{y}_{\text{obs}} - \mathbf{x}^{*(t)}_{\text{obs}}$, and $\mathbf{r}_{\text{obs}} = \mathbf{d} - \boldsymbol{\mu}_{\mathbf{Z}^*_{\text{obs}}}$ is its residual with respect to the prior mean. The correction over the entire grid is given by the MAP estimate, i.e., the GP predictive mean conditioned on $\mathbf{d}$:
\begin{align}
\hat{\boldsymbol{\mu}}_{\mathbf{Z}^*} = \boldsymbol{\mu}_{\mathbf{Z}^*} + \mathbf{K}_{\boldsymbol{\gamma},\mathbf{Z}^*,\mathbf{Z}^*_{\text{obs}}} \mathbf{K}_{\mathbf{d}}^{-1} \mathbf{r}_{\text{obs}},
\label{eq:mu_S}
\end{align}
where $\mathbf{K}_{\mathbf{d}} = \mathbf{K}_{\boldsymbol{\gamma},\mathbf{Z}^*_{\text{obs}},\mathbf{Z}^*_{\text{obs}}} + \sigma^2 \mathbf{I}_m \in \mathbb{R}^{m \times m}$ is the covariance of the departures and $\mathbf{K}_{\boldsymbol{\gamma},\mathbf{Z}^*,\mathbf{Z}^*_{\text{obs}}} \in \mathbb{R}^{HW \times m}$ the cross-covariance between all grid points and the observed ones. The assimilated field $\hat{\mathbf{y}} \in \mathbb{R}^{HW}$ is obtained by adding this correction to the background:
\begin{align}
\hat{\mathbf{y}} = \mathbf{x}^{*(t)} + \hat{\boldsymbol{\mu}}_{\mathbf{Z}^*}.
\label{eq:analysis}
\end{align}
Owing to the noise term with covariance $\sigma^2 \mathbf{I}_m$, the noisy measurements are smoothed at the observed points, while the background is corrected at all unobserved ones. The uncertainty of the correction is quantified by the GP predictive covariance
\begin{align}
\hat{\boldsymbol{\Sigma}}_{\mathbf{Z}^*} = \mathbf{K}_{\boldsymbol{\gamma},\mathbf{Z}^*,\mathbf{Z}^*} - \mathbf{K}_{\boldsymbol{\gamma},\mathbf{Z}^*,\mathbf{Z}^*_{\text{obs}}} \mathbf{K}_{\mathbf{d}}^{-1} \mathbf{K}_{\boldsymbol{\gamma},\mathbf{Z}^*_{\text{obs}},\mathbf{Z}^*},
\end{align}
which provides a principled measure of the reliability of the assimilated state.

To turn this uncertainty into an adaptive correction strength, we introduce a modulation parameter $\alpha \geq 0$. Let $$\hat{\boldsymbol{\sigma}}_{\mathbf{Z}^*} = \sqrt{\operatorname{diag}(\hat{\boldsymbol{\Sigma}}_{\mathbf{Z}^*})} \in \mathbb{R}^{HW}$$ be the pointwise predictive standard deviation. The correction is rescaled by the weight $\mathbf{w}_\alpha = \exp(-\alpha \, \hat{\boldsymbol{\sigma}}_{\mathbf{Z}^*})$, with the square root and the exponential applied elementwise, yielding the modulated correction $\tilde{\boldsymbol{\mu}}_{\mathbf{Z}^*} = \mathbf{w}_\alpha \odot \hat{\boldsymbol{\mu}}_{\mathbf{Z}^*}$, which replaces $\hat{\boldsymbol{\mu}}_{\mathbf{Z}^*}$ in \eqref{eq:analysis} when the modulation is enabled. For $\alpha \to 0$, $\mathbf{w}_\alpha \to \mathbf{1}$ and the unmodulated correction is recovered. As $\alpha$ increases, the correction is increasingly concentrated in low-uncertainty regions and damped in poorly observed ones, providing a simple, interpretable trade-off between assimilation strength and confidence. The inference procedure is summarized in Algorithm \ref{alg:inference}.

\begin{algorithm}
\caption{Inference and data assimilation}
\label{alg:inference}
\begin{algorithmic}[1]
\Require Background state $\mathbf{x}^*$, observations $\mathbf{y}_{\text{obs}} \in \mathbb{R}^m$, trained $\boldsymbol{\theta}$, $\boldsymbol{\gamma}$, $\sigma^2$, modulation parameter $\alpha$
\Ensure Assimilated field $\hat{\mathbf{y}}$, predictive covariance $\hat{\boldsymbol{\Sigma}}_{\mathbf{Z}^*}$
\State $\mathbf{Z}^* \gets g_{\boldsymbol{\theta}}(\mathbf{x}^*)$ \Comment{$\mathbf{Z}^* \in \mathbb{R}^{h \times HW}$}
\State $\mathbf{Z}^*_{\text{obs}} \gets$ columns of $\mathbf{Z}^*$ at observed points
\State $\mathbf{d} \gets \mathbf{y}_{\text{obs}} - \mathbf{x}^{*(t)}_{\text{obs}}$ \Comment{departures}
\State $\mathbf{K}_{\boldsymbol{\gamma},\mathbf{Z}^*_{\text{obs}},\mathbf{Z}^*_{\text{obs}}} \gets [k_{\boldsymbol{\gamma}}(\mathbf{z}_i, \mathbf{z}_j)]_{i,j}$
\State $\boldsymbol{\mu}_{\mathbf{Z}^*_{\text{obs}}} \gets$ prior mean at $\mathbf{Z}^*_{\text{obs}}$
\State $\mathbf{K}_{\mathbf{d}} \gets \mathbf{K}_{\boldsymbol{\gamma},\mathbf{Z}^*_{\text{obs}},\mathbf{Z}^*_{\text{obs}}} + \sigma^2 \mathbf{I}_m$
\State $\boldsymbol{\beta} \gets \mathbf{K}_{\mathbf{d}}^{-1} (\mathbf{d} - \boldsymbol{\mu}_{\mathbf{Z}^*_{\text{obs}}})$
\For{each patch $k$}
    \State $\mathbf{Z}^*_k \gets$ patch $k$ of $\mathbf{Z}^*$
    \State $\mathbf{K}_{\boldsymbol{\gamma},\mathbf{Z}^*_k,\mathbf{Z}^*_{\text{obs}}} \gets [k_{\boldsymbol{\gamma}}(\mathbf{z}_i, \mathbf{z}_j)]_{i,j}$
    \State $\boldsymbol{\mu}_{\mathbf{Z}^*_k} \gets$ prior mean at $\mathbf{Z}^*_k$
    \State $\hat{\boldsymbol{\mu}}_{\mathbf{Z}^*_k} \gets \boldsymbol{\mu}_{\mathbf{Z}^*_k} + \mathbf{K}_{\boldsymbol{\gamma},\mathbf{Z}^*_k,\mathbf{Z}^*_{\text{obs}}}\, \boldsymbol{\beta}$
\EndFor
\State Stack $\{\hat{\boldsymbol{\mu}}_{\mathbf{Z}^*_k}\}$ into $\hat{\boldsymbol{\mu}}_{\mathbf{Z}^*}$
\State Stack $\{\mathbf{K}_{\boldsymbol{\gamma},\mathbf{Z}^*_k,\mathbf{Z}^*_{\text{obs}}}\}$ into $\mathbf{K}_{\boldsymbol{\gamma},\mathbf{Z}^*,\mathbf{Z}^*_{\text{obs}}}$
\State $\mathbf{K}_{\boldsymbol{\gamma},\mathbf{Z}^*,\mathbf{Z}^*} \gets [k_{\boldsymbol{\gamma}}(\mathbf{z}_i, \mathbf{z}_j)]_{i,j}$
\State $\mathbf{V} \gets \mathbf{K}_{\mathbf{d}}^{-1} \mathbf{K}_{\boldsymbol{\gamma},\mathbf{Z}^*,\mathbf{Z}^*_{\text{obs}}}^\top$
\State $\hat{\boldsymbol{\Sigma}}_{\mathbf{Z}^*} \gets \mathbf{K}_{\boldsymbol{\gamma},\mathbf{Z}^*,\mathbf{Z}^*} - \mathbf{K}_{\boldsymbol{\gamma},\mathbf{Z}^*,\mathbf{Z}^*_{\text{obs}}}\, \mathbf{V}$
\State $\mathbf{w}_\alpha \gets \exp\big(-\alpha \sqrt{\operatorname{diag}(\hat{\boldsymbol{\Sigma}}_{\mathbf{Z}^*})}\big)$
\State $\hat{\mathbf{y}} \gets \mathbf{x}^{*(t)} + \mathbf{w}_\alpha \odot \hat{\boldsymbol{\mu}}_{\mathbf{Z}^*}$
\State \Return $\hat{\mathbf{y}}$, $\hat{\boldsymbol{\Sigma}}_{\mathbf{Z}^*}$
\end{algorithmic}
\end{algorithm}

\subsection{Scalable and Memory-Efficient Approximation through Spatial Patching}
\label{subsec:patching}

Evaluating the GP over the whole domain requires building the dense matrix $\mathbf{K}_{\boldsymbol{\gamma},\mathbf{Z},\mathbf{Z}} \in \mathbb{R}^{HW \times HW}$, factorizing it via Cholesky decomposition and computing its log-determinant, with time and memory complexities of $\mathcal{O}((HW)^3)$ and $\mathcal{O}((HW)^2)$, respectively. To reduce this cost, we adopt a spatial patching strategy.

During training, the grid is partitioned into $K$ non-overlapping square patches of $P \times P$ points ($P \ll H, W$), which are treated as samples within mini-batches for Stochastic Gradient Descent (SGD). For the $k$-th patch, let $\mathbf{Z}_k \in \mathbb{R}^{h \times P^2}$ collect the columns of $\mathbf{Z}$ at its grid points, and $\mathbf{y}_k \in \mathbb{R}^{P^2}$ and $\boldsymbol{\mu}_{\mathbf{Z}_k} \in \mathbb{R}^{P^2}$ the corresponding entries of the background error and of the prior mean. The NMLL is then evaluated locally as
\begin{align}
\mathcal{L}_{\text{NMLL}}^{(k)} = \tfrac{1}{2}\, \mathbf{r}_k^\top \mathbf{K}_{\mathbf{y}_k}^{-1} \mathbf{r}_k + \tfrac{1}{2} \log |\mathbf{K}_{\mathbf{y}_k}|,
\end{align}
where $\mathbf{r}_k = \mathbf{y}_k - \boldsymbol{\mu}_{\mathbf{Z}_k}$ and $\mathbf{K}_{\mathbf{y}_k} = \mathbf{K}_{\boldsymbol{\gamma},\mathbf{Z}_k,\mathbf{Z}_k} + \sigma^2 \mathbf{I}_{P^2}$. Each patch costs $\mathcal{O}(P^6)$ and $HW/P^2$ patches cover the domain, so the training cost per sample scales as $\mathcal{O}(HW P^4)$: for a fixed patch size it grows linearly with the number of grid points rather than cubically, making the approach applicable to large domains.

At inference, instead, patch-wise evaluation is exact rather than approximate. Let $\mathbf{Z}^*_k \in \mathbb{R}^{h \times P^2}$ collect the columns of $\mathbf{Z}^*$ in the $k$-th patch. By the linearity of the GP predictive mean, the correction restricted to each patch is
\begin{align}
\hat{\boldsymbol{\mu}}_{\mathbf{Z}^*_k} = \boldsymbol{\mu}_{\mathbf{Z}^*_k} + \mathbf{K}_{\boldsymbol{\gamma},\mathbf{Z}^*_k,\mathbf{Z}^*_{\text{obs}}} \boldsymbol{\beta}, \qquad \boldsymbol{\beta} = \mathbf{K}_{\mathbf{d}}^{-1} \mathbf{r}_{\text{obs}}.
\end{align}
Since $\boldsymbol{\beta} \in \mathbb{R}^m$ depends only on the observed points, it is computed once, and the full correction is obtained by stacking the local predictions, $\hat{\boldsymbol{\mu}}_{\mathbf{Z}^*} = [\hat{\boldsymbol{\mu}}_{\mathbf{Z}^*_1}^\top, \dots, \hat{\boldsymbol{\mu}}_{\mathbf{Z}^*_K}^\top]^\top$, up to a reordering of the grid points. This yields exactly the same result as a full-domain pass, while bounding the memory required on large outputs.

\section{Experimental Setup and Results}
\label{experimental_results}

\begin{figure}
    \centering
    \includegraphics[width=\linewidth]{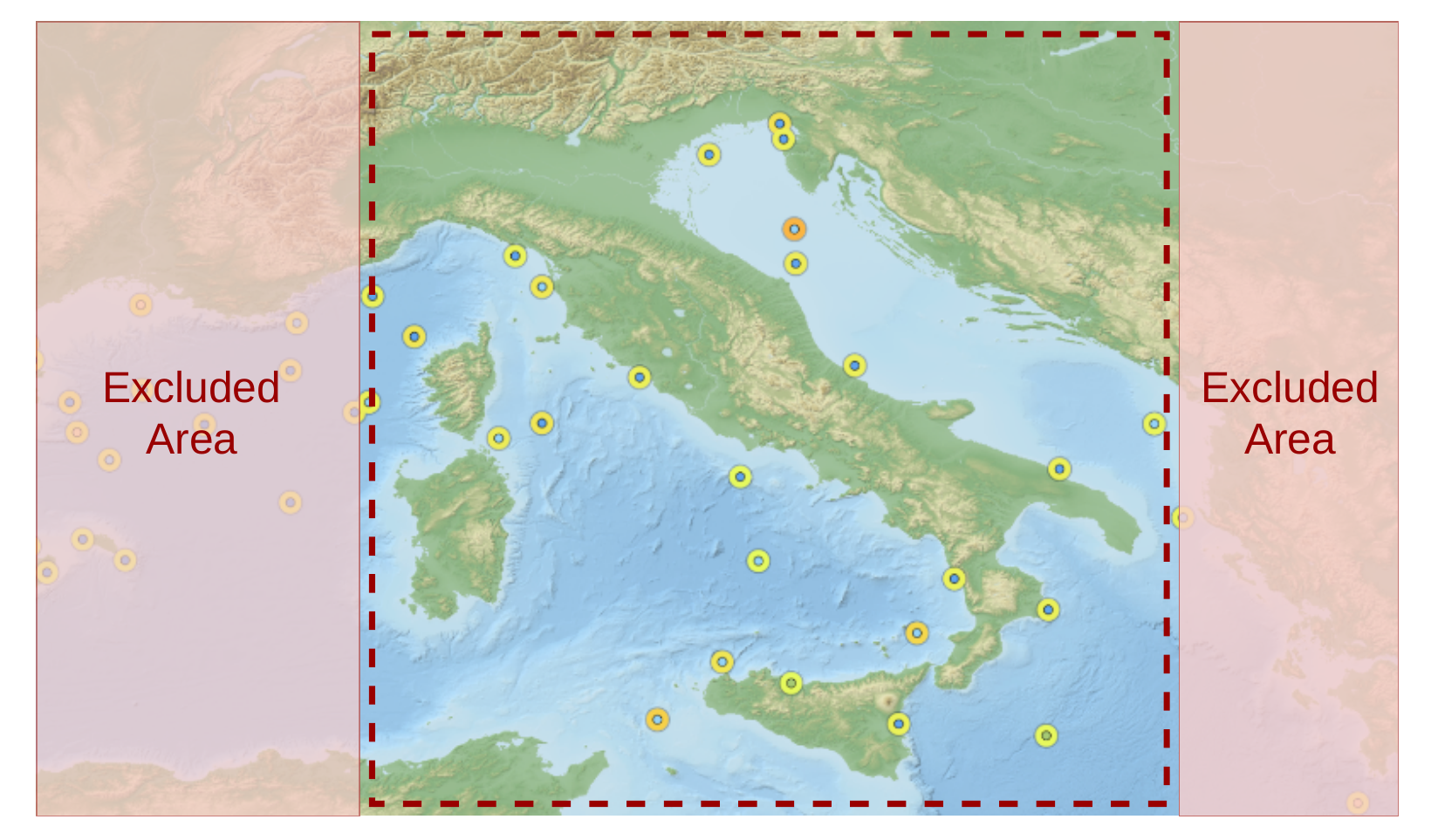}
    \caption{Mediterranean Sea domain selected for the numerical experiments. The dashed box delimits the $128 \times 128$ computational domain; the shaded areas outside it are excluded. Yellow markers indicate the positions of in-situ mooring buoys.}    
    \label{fig:italy}
\end{figure}

To evaluate the performance of the proposed approach, the assimilation technique is tested within the surface of the Mediterranean Sea, particularly focusing on the marine sectors surrounding the Italian peninsula, as shown in Figure~\ref{fig:italy}. Particular focus is given to the buoys' observations since they are the most reliable and, at the same time, challenging to assimilate because of the scarcity of the available data.
 In particular, two experiments are performed:
\begin{itemize}
	\item the assimilation of noisy synthetic observations sampled from the ground truth, to evaluate the framework as the observation density grows;
	\item the assimilation of real, extremely sparse in-situ observations, to assess its behavior in an operational setting.
\end{itemize}
To ensure a rigorous comparison, the proposed method is evaluated against three main models. 

First, a classical GP model is implemented as a baseline. Since the GP approach entirely discards the CNN backbone while retaining the same assimilation framework, it also serves as an ablation study to isolate and quantify the specific contribution of the CNN; its formulation follows directly from the GP framework introduced in Section \ref{background}.

Second, the proposed method is compared against the consolidated 3D-Var algorithm. In classic 3D-Var data assimilation, the background state $\mathbf{x}_b \in \mathbb{R}^n$, with $n$ the number of ocean grid points, is corrected using the observations $\mathbf{y}_{\text{obs}} \in \mathbb{R}^m$ by minimizing the variational cost function:
\begin{align}
J(\mathbf{x}) = {} & \tfrac{1}{2} (\mathbf{x} - \mathbf{x}_b)^\top \mathbf{B}^{-1} (\mathbf{x} - \mathbf{x}_b) \nonumber \\
& + \tfrac{1}{2} (\mathbf{y}_{\text{obs}} - \mathbf{H}\mathbf{x})^\top \mathbf{R}^{-1} (\mathbf{y}_{\text{obs}} - \mathbf{H}\mathbf{x}),
\label{eq:3dvar_cost}
\end{align}
where $\mathbf{B} \in \mathbb{R}^{n \times n}$ and $\mathbf{R} \in \mathbb{R}^{m \times m}$ represent the background and observation error covariance matrices, respectively, and $\mathbf{H} \in \mathbb{R}^{m \times n}$ is the linear observation operator. The assimilated state can be computed as $\mathbf{x}_a = \arg\min_{\mathbf{x}} J(\mathbf{x})$.
To ensure a fair comparison, the covariance matrices $\mathbf{B}$ and $\mathbf{R}$ are constructed using the same kernel family as our proposed hybrid framework. Furthermore, their hyperparameters are fine-tuned via Bayesian Optimization to directly minimize the Root Mean Squared Error (RMSE) on the training and validation dataset.
Additionally, since the spatial domain is defined over a 2D image grid, full domain covariance matrices $\mathbf{K}_{\text{full}} = \mathbf{K}_{\text{lon}} \otimes \mathbf{K}_{\text{lat}}$ are initially computed using Kronecker products ($\otimes$) along the spatial dimensions. These are subsequently masked/sliced to remove land grid points, ensuring that all error covariance operators $\mathbf{B}$ and $\mathbf{R}$ strictly operate over valid ocean locations.

Third, we developed a purely deep-learning competitor, referred to as FiLM CNN, which shares the identical convolutional backbone of the proposed method but replaces the GP correction stage with a direct regression head; sparse observations are injected both as additional input channels and, at the latent-embedding level, through Feature-wise Linear Modulation (FiLM) \cite{perez2018film}, a general-purpose conditioning mechanism originally proposed for visual reasoning. Complementary to the GP-only ablation, the FiLM CNN isolates the specific contribution of the GP layer, since it retains the same neural backbone and observation-conditioning philosophy as the proposed method but lacks its probabilistic, kernel-based mechanism for propagating information from the sparse observations across the domain.

Finally, since both 3D-Var and the classical GP assume an unbiased background error, while the proposed method and the FiLM CNN learn the systematic bias implicitly from data, a direct comparison would penalize the former for a limitation that is routinely addressed in practice. We therefore also test an offline climatological bias correction for 3D-Var and the GP, as commonly done in operational systems: the mean background error is estimated per grid point and calendar month over the training and validation periods only, without using the test year, and is added to the background before the assimilation. For SSS, this correction substantially improves both competitors and is therefore adopted in all SSS experiments. For SST, instead, it slightly degrades their accuracy, suggesting that the SST background error is dominated by a state-dependent rather than a recurring seasonal component; to compare against the strongest version of each competitor, the correction is therefore not applied to SST.

\subsection{Datasets}

To ensure reproducibility, we evaluate the framework on publicly available oceanographic datasets, considering four variables: SST, SSS, and the zonal and meridional components of Sea Surface Velocity (SSU and SSV). All four fields are provided as inputs to the feature extractor, while DA is performed on SST and SSS.

The GLORYS12V1 reanalysis\footnote{\url{https://data.marine.copernicus.eu/product/GLOBAL\_MULTIYEAR\_PHY\_001\_030/description}} is used as the ground truth $\mathbf{y}_{\text{gt}}$, since reanalyses provide the most reliable reference by integrating multi-source observations with physical models. The background state $\mathbf{x}$, concatenating the four variables, is taken from the uncorrected forecasts of the Operational Mercator global ocean analysis and forecast system\footnote{\url{https://data.marine.copernicus.eu/product/GLOBAL\_ANALYSISFORECAST\_PHY\_001\_024/description}}, reproducing realistic operational conditions. Both products have a native resolution of $1/12^\circ$ and are extracted over $[8.02^\circ\text{E}, 18.68^\circ\text{E}] \times [35.75^\circ\text{N}, 46.35^\circ\text{N}]$ (approximately $1.03 \times 10^6\,\text{km}^2$) at a depth of about $0.49$\,m. This yields an $H \times W = 128 \times 128$ grid with $9{,}709$ ocean points out of $16{,}384$. Although the reanalysis spans several decades, the operational forecasts are only available over a much shorter period, which therefore bounds the number of background--ground truth pairs that can be used for training and evaluation. The daily dataset covers the entire period in which both products are available, from June 1, 2022 to April 1, 2026: the period from June 2022 to June 2024 is used for training, the year 2025 is reserved as the independent test set, and July--December 2024 together with January--March 2026 are used as validation set for model selection and for the choice of $\alpha$.

Sparse in-situ observations $\mathbf{y}_\text{obs}$ are taken from the \mbox{INSTAC} product\footnote{\url{https://data.marine.copernicus.eu/product/INSITU\_MED\_PHYBGCWAV\_DISCRETE\_MYNRT\_013\_035/description}}. To avoid low-quality data, we retain SST measurements with quality level $1$ and, given their scarcity, SSS measurements with levels $1$ to $3$. Days without valid measurements are excluded from the assimilation. Although GLORYS12V1 also assimilates in-situ measurements, its surface fields are mainly constrained by satellite observations, such as sea surface temperature and altimetry, which provide a much denser coverage; the contribution of the sparse in-situ buoys to the reanalysis is therefore marginal, limiting their dependence on the ground truth.

\subsection{Neural Network \& Gaussian Process Setup}
\label{subsec:nn_gp_setup}

\begin{figure}
    \centering
    \includegraphics[width=\linewidth]{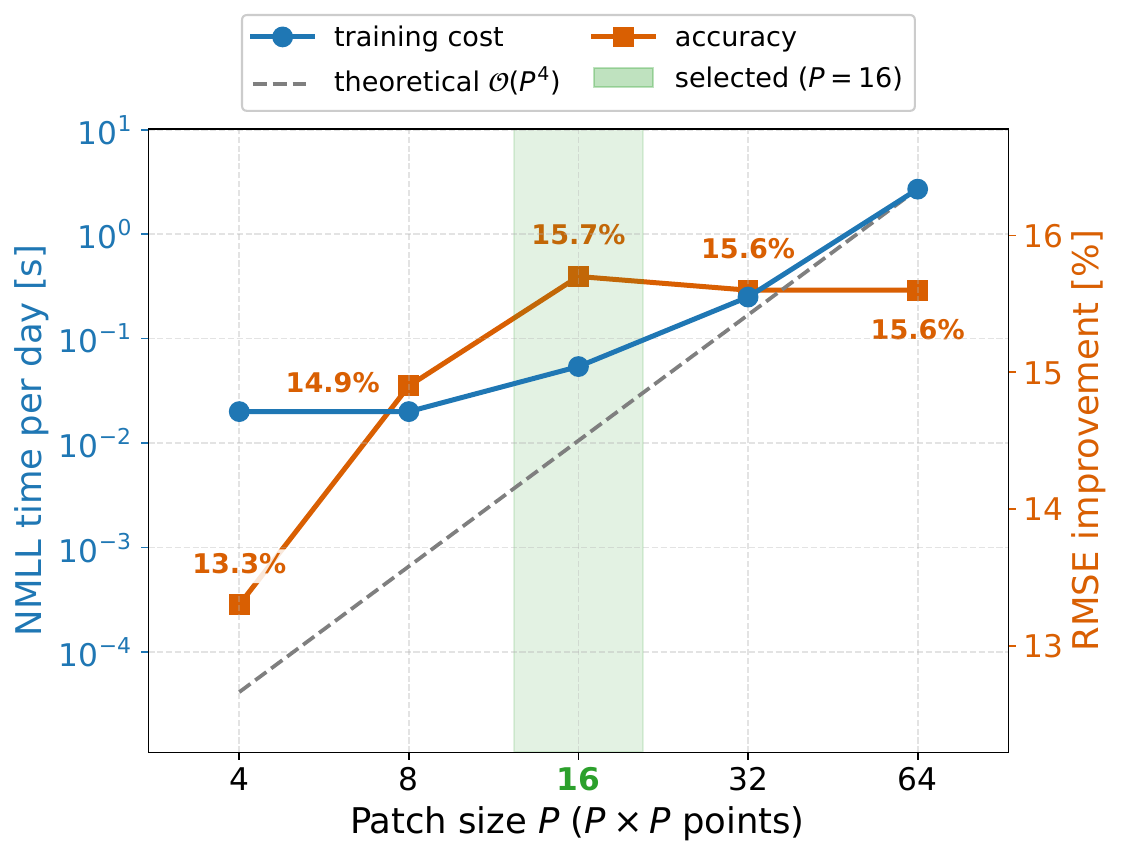}
    \caption{Training cost per day and RMSE improvement over the unassimilated simulation as a function of the patch size.}
    \label{fig:patch_cost_accuracy}
\end{figure}

The feature extractor is a lightweight CNN with four convolutional layers and $304{,}384$ trainable parameters. The first three layers follow a U-Net-style encoder-decoder with a $64$-dimensional bottleneck, while the last one projects the features into a latent space of dimension $h=32$, preserving the spatial resolution of the input. Four auxiliary channels are concatenated with the input to provide spatial and temporal context: a binary land-sea mask (equal to $0$ on land), two channels with the latitude and longitude of each grid point, and a sinusoidal time embedding encoding month and year. The resulting embedding matrix $\mathbf{Z}$ provides, at each grid point $i$, a latent vector $\mathbf{z}_i$ encoding inter-variable dependencies.

The GP uses an RBF kernel with Automatic Relevance Determination (ARD):
\begin{equation}
k_{\boldsymbol{\gamma}}(\mathbf{z}_i, \mathbf{z}_j) = s^2 \exp\left( -\frac{1}{2} \sum_{l=1}^{h} \frac{(z_{i,l} - z_{j,l})^2}{\ell_l^2} \right),
\end{equation}
where $z_{i,l}$ is the $l$-th entry of $\mathbf{z}_i$, $s^2$ is the outputscale and $\{\ell_l\}_{l=1}^{h}$ are feature-specific lengthscales that weight the relevance of each latent dimension. The GP has $68$ trainable hyperparameters: $33$ for the linear prior mean $\boldsymbol{\mu}_{\mathbf{Z}} = \mathbf{Z}^\top \mathbf{w} + w_0 \mathbf{1}_{HW}$, with $\mathbf{w} \in \mathbb{R}^{h}$ and $w_0 \in \mathbb{R}$; $33$ for the kernel ($h$ lengthscales and $s^2$); and $2$ for the Gaussian likelihood.
A patch size of $P = 16$ is used, selected through a grid search hyper-parameter evaluation with $P \in \{4, 8, 16, 32, 64\}$ in which the proposed framework is re-trained and evaluated on the validation set: results reported in Figure~\ref{fig:patch_cost_accuracy} show that smaller patches reduce the accuracy, while larger ones increase the training cost without further improvements. For small patches, the measured cost exceeds the $\mathcal{O}(HWP^4)$ scaling, as it is dominated by the overhead of processing many small patches.

The value of $\alpha$ is selected on the validation set for each variable and observation density; its effect is analyzed in Section~\ref{numerical_results}.

The architecture is trained end-to-end for $500$ epochs with constant learning rates of $10^{-4}$ for the feature extractor and $10^{-3}$ for the GP hyperparameters. To facilitate reproducibility, the source code will be released publicly with the final version of the paper.

\subsection{Numerical results}
\label{numerical_results}


The first experiment aims to evaluate the impact of the volume of the available observations on data assimilation performance. Because real-world measurements are limited, model behavior across varying observation densities is evaluated using noisy synthetic observations. Performance is measured by the mean daily RMSE, i.e., the RMSE over all ocean points computed for each day and averaged over the test year; observation locations are kept fixed across days to isolate the impact of observation density.

\begin{figure*}
    \centering
    \includegraphics[width=\linewidth]{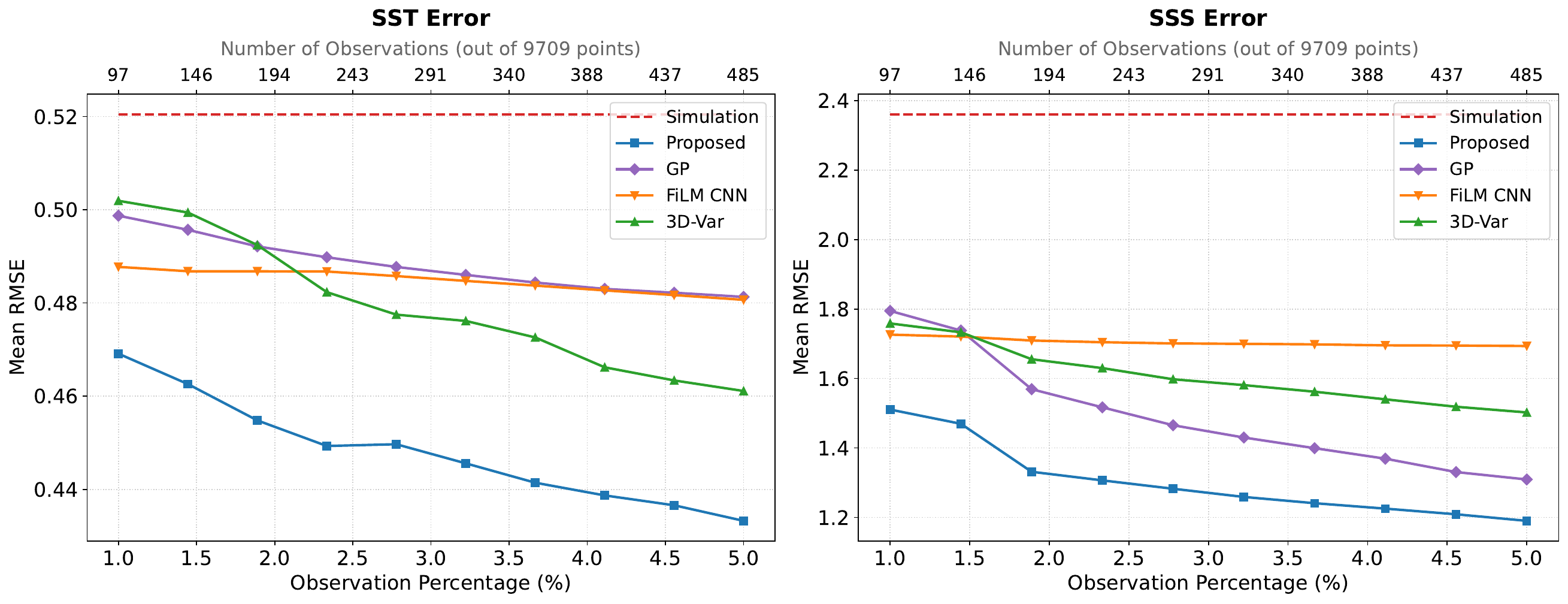}
    \caption{Mean daily RMSE of SST and SSS over the test year as a function of the density of synthetic observations. The dashed line is the unassimilated simulation.}
    \label{fig:eval}
\end{figure*}

Figure \ref{fig:eval} (left panel) compares the SST mean RMSE of the proposed approach and the competing algorithms against the unassimilated baseline (dashed line). Our model achieves the lowest RMSE at all observation ratios, remaining robust even under extreme sparsity. At $1\%$ observation density ($97$ observations out of $9{,}709$ spatial points), it improves on the baseline by approximately $10\%$, against $6\%$ for the FiLM CNN and $4\%$ for both \mbox{3D-Var} and the GP. At $5\%$ observation density the margin widens, with approximately $16\%$ for our model versus $12\%$ for 3D-Var and $7\%$ for both the GP and the FiLM CNN. The gains of the proposed method and \mbox{3D-Var} scale linearly with the number of observations, whereas those of the GP grow more slowly. At low densities, the FiLM CNN outperforms both model-based competitors thanks to a globally learned bias correction applied regardless of the observation locations. However, it improves only marginally as observations increase, and is eventually overtaken by \mbox{3D-Var} and matched by the GP.

For SSS in Figure~\ref{fig:eval} (right panel) the climatological bias correction substantially reduces the error of 3D-Var and the GP, which already lie well below the unassimilated simulation at the lowest density. Nevertheless, the proposed method achieves the lowest RMSE at all observation ratios. At $1\%$, it reduces the baseline error by approximately $36\%$, against $27\%$ for the FiLM CNN, $25\%$ for 3D-Var and $24\%$ for the GP. At $5\%$, the reduction reaches approximately $50\%$, versus $44\%$ for the GP, $36\%$ for 3D-Var and $28\%$ for the FiLM CNN. At the lowest densities the three competitors perform similarly, since their gains are dominated by bias removal, climatological for 3D-Var and the GP and learned for the FiLM CNN, rather than by the observations. As the density increases, 3D-Var and the GP progressively benefit from local conditioning, with the GP overtaking 3D-Var beyond $1.5\%$, whereas the FiLM CNN remains almost flat. The error of the proposed method decreases monotonically, with the largest gain between $1.4\%$ and $1.9\%$. Its margin over the bias-corrected competitors indicates that the learned, state-dependent bias term captures more than the climatological mean bias.

\begin{figure*}
    \centering
    \includegraphics[width=\linewidth]{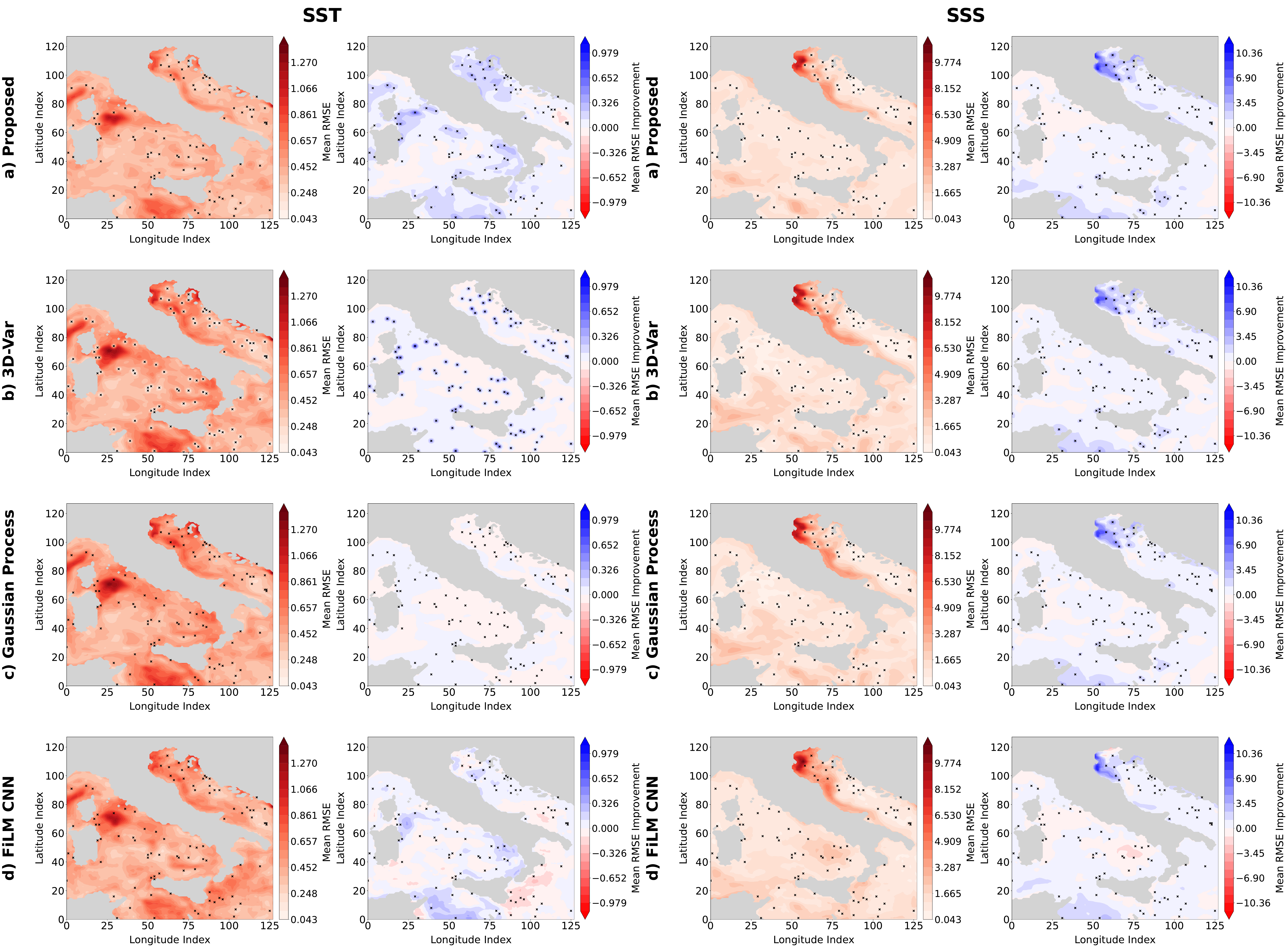}
    \caption{Spatial distribution of the SST and SSS assimilation error and of the error reduction with respect to the unassimilated baseline. Black markers indicate the observation locations.}
    \label{fig:spatial_comparison}
\end{figure*}

Figure~\ref{fig:spatial_comparison} shows, for both variables with $1\%$ of synthetic observations, the spatial distribution of the error and of the error reduction with respect to the unassimilated baseline. The corresponding mean daily RMSE values are those reported in Figure~\ref{fig:eval} at $1\%$, so this figure is meant as a qualitative illustration of the spatial structure of the corrections. For SST, the proposed method produces broad, spatially coherent improvements that extend well beyond the observation points, whereas 3D-Var yields isotropic corrections centered on each observation, as imposed by its Gaussian background error covariance $\mathbf{B}$. The GP achieves only marginal improvements, while the FiLM CNN produces a diffuse error reduction only weakly tied to the observation locations. For SSS, the largest errors and corrections of all methods are concentrated in the northern Adriatic, along the Po River plume, where the proposed method achieves the strongest error reduction.

\begin{figure}
    \centering
    \includegraphics[width=\linewidth]{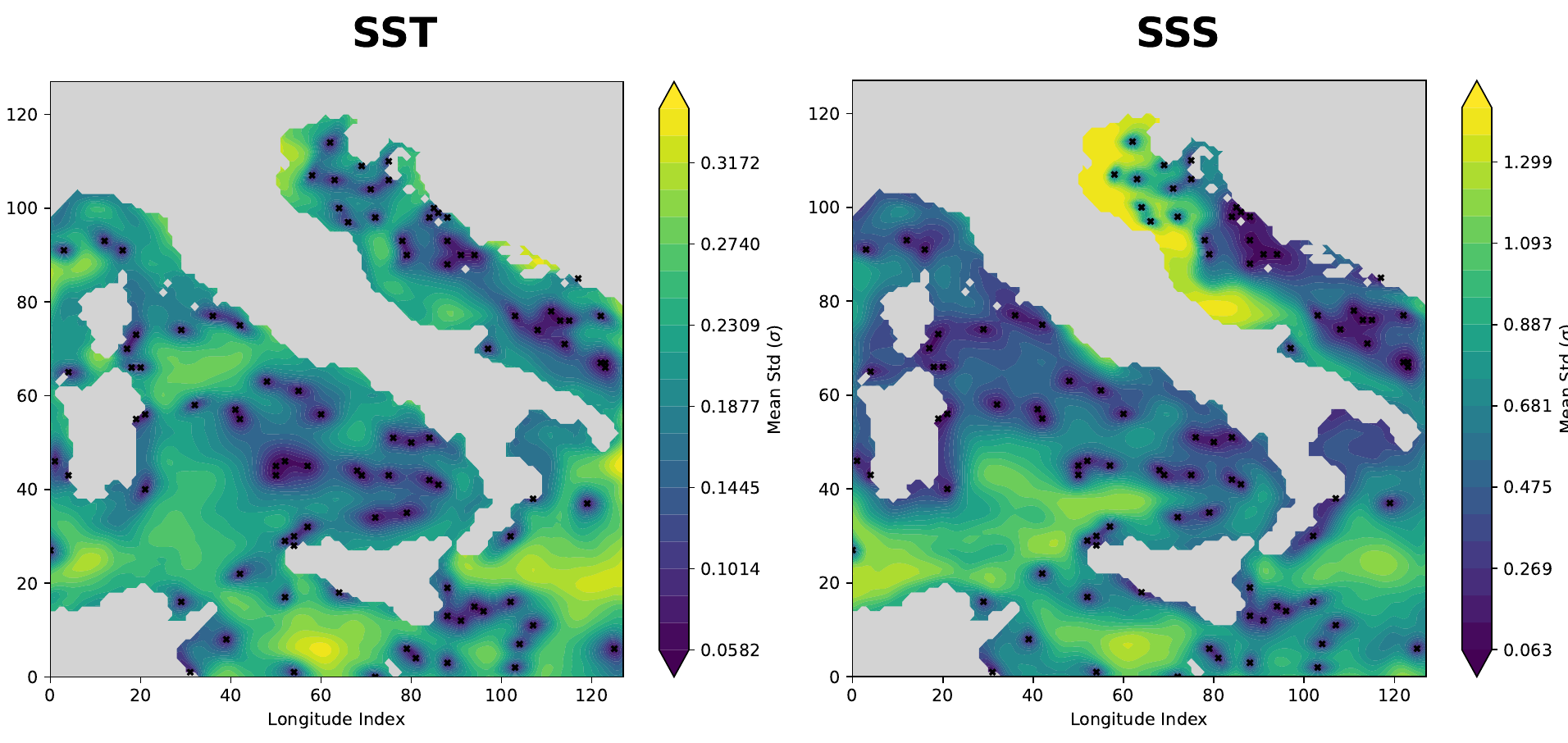}
    \caption{Posterior standard deviation estimated by the proposed model over the test year for SST and SSS. Black markers indicate the observation locations.}
    \label{fig:spatial_std}
\end{figure}

Furthermore, the proposed framework provides an explicit measure of predictive uncertainty through spatial standard deviation maps, as shown in Figure \ref{fig:spatial_std}, obtained in the same setting as Figure~\ref{fig:spatial_comparison}. Owing to the GP structure, for both variables the standard deviation is lower around the observations and higher in unobserved regions, reflecting the model's confidence in its corrections. For SSS, the highest uncertainty is concentrated in the northern Adriatic along the Po River plume, even in the presence of nearby observations: this is the region where Figure~\ref{fig:spatial_comparison} shows the largest background errors, so the predictive variance also flags where the corrections are least reliable.

\begin{figure*}
    \centering
    \includegraphics[width=\linewidth]{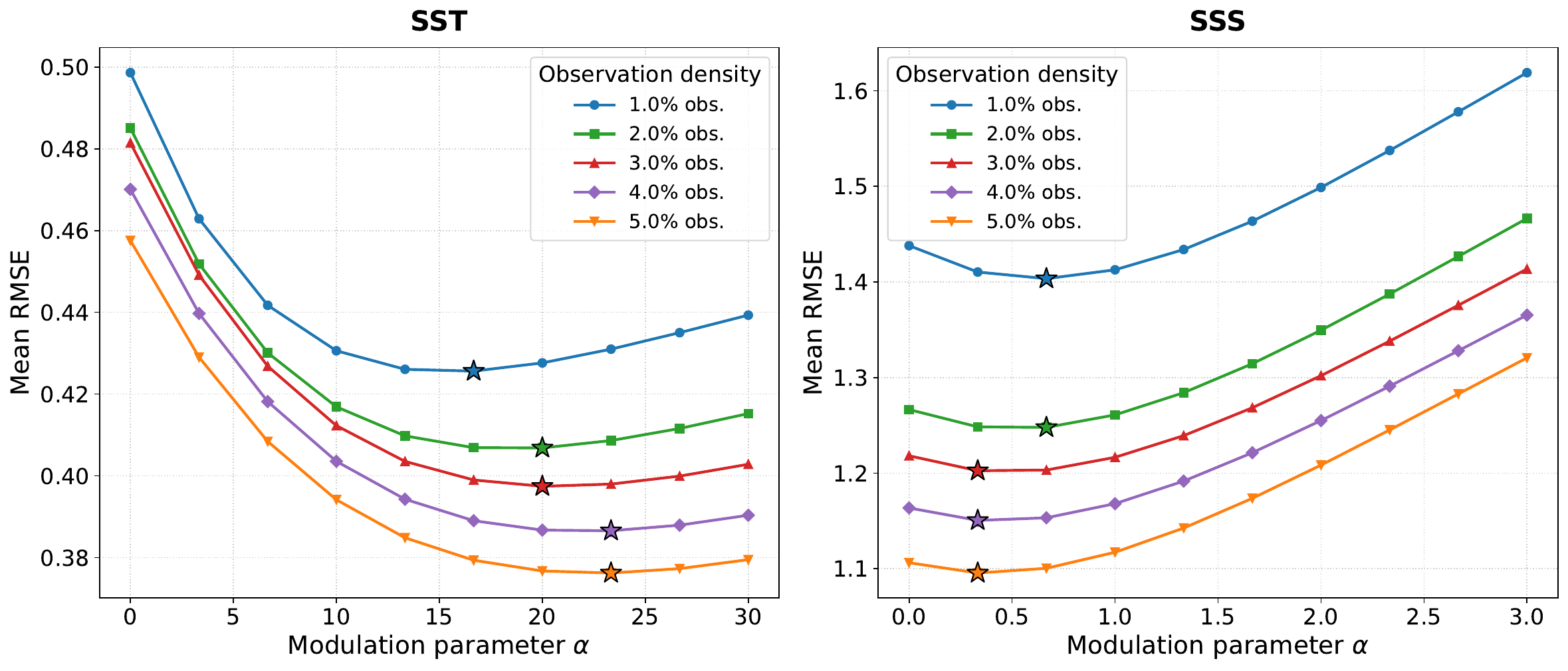}
    \caption{Mean daily RMSE of the proposed method as a function of $\alpha$ for SST and SSS, at different densities of synthetic observations. Stars mark the best value of $\alpha$ for each density on the test year.}
\label{fig:alpha_sensitivity_combined}
\end{figure*}

Building on this notion of predictive uncertainty, the modulation parameter $\alpha$ strengthens corrections in confident regions and dampens them in high-uncertainty areas, preventing over-correction driven by simulation biases. Figure~\ref{fig:alpha_sensitivity_combined} shows the sensitivity of the assimilation accuracy to $\alpha$ at five densities of synthetic observations, from $1\%$ to $5\%$. For SST, the modulation is clearly beneficial: starting from the unmodulated case ($\alpha=0$), the RMSE drops markedly and reaches a broad minimum, from $\alpha \approx 17$ at $1\%$ to $\alpha \approx 23$ at $4\%$ and $5\%$, after which it slightly increases. At the optimum, the error is reduced by approximately $15\%$ at $1\%$ and $18\%$ at $5\%$ with respect to $\alpha=0$. The optimal $\alpha$ grows with the observation density: as more observations lower $\hat{\boldsymbol{\sigma}}_{\mathbf{Z}^*}$ over a larger part of the domain, a larger $\alpha$ is needed to obtain a comparable damping in the regions that remain poorly observed.

For SSS, all curves share a similar shape, with a shallow minimum at small values, $\alpha \approx 0.3$--$0.7$, followed by a steady increase, and the gain with respect to $\alpha=0$ is limited to approximately $1$--$2\%$. This is consistent with a correction dominated by the learned bias term, which spans the whole basin. Note that the results in Figure~\ref{fig:eval} are obtained with the value of $\alpha$ selected on the validation set, which does not necessarily coincide with the test year optimum marked by the stars in Figure~\ref{fig:alpha_sensitivity_combined}. Overall, the optimal $\alpha$ depends on both the variable and the observation density, making the assimilation strength a flexible, tunable design choice.

We evaluate the model in an operational setting by performing DA over the test year with real buoy observations, with $\alpha$ selected on the validation set as in the synthetic experiments. High-quality observations are available on only 189 days for SST and 125 days for SSS, and the number and positions of the buoys change over time. At most, 6 SST and 3 SSS observations out of $9{,}709$ points are available in a single day ($\approx 0.06\%$ and $\approx 0.03\%$, respectively), with fewer on average.

\begin{figure}
    \centering
    \includegraphics[width=\linewidth]{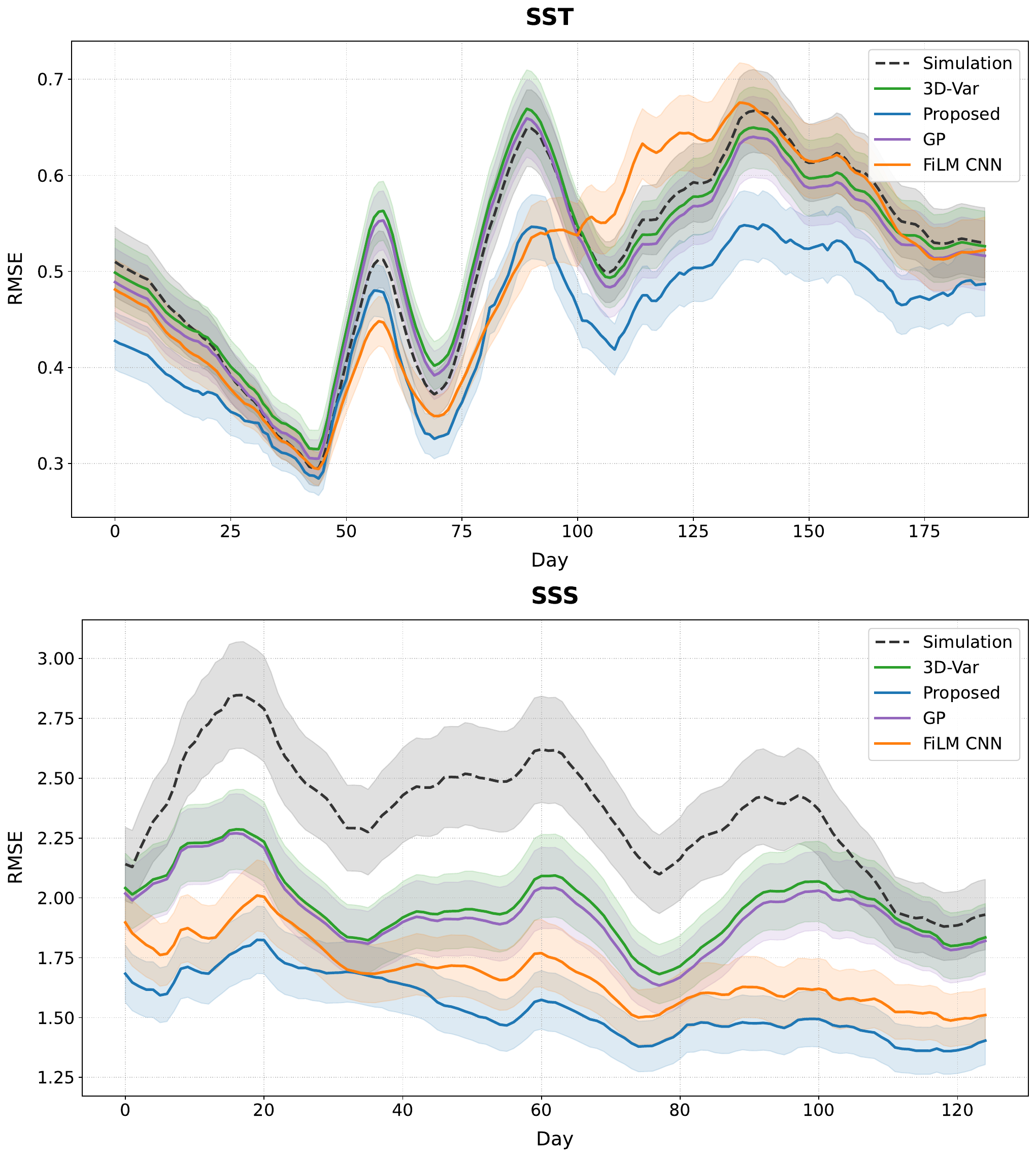}
    \caption{Daily mean RMSE of SST and SSS over the test year with real buoy observations.}    
    \label{fig:time_rmse_real}
\end{figure}

For SST in Figure \ref{fig:time_rmse_real} (top panel), the proposed model overcomes this scarcity by simultaneously assimilating the observations and correcting the bias, remaining below all competitors for most of the period. 3D-Var and the GP improve only slightly on the baseline: lacking bias correction, their observation-driven updates are confined to the areas covered by the observations. The FiLM CNN network performs competitively for roughly the first half of the period, and in some cases matches the proposed method. It then degrades markedly, exceeding even the unassimilated simulation (approximately between the 100th and the 170th available day), before converging again toward the end.

As to SSS in Figure \ref{fig:time_rmse_real} (bottom panel), this is more challenging since approximately $0.03\%$ of observations are barely sufficient to affect the assimilated state. Here, the climatological bias correction brings 3D-Var and the GP well below the unassimilated simulation, with two nearly overlapping curves. Given the extreme sparsity, their gain comes almost entirely from the removal of the mean seasonal bias, and it narrows toward the end of the period. The proposed method achieves the lowest error for almost the entire period, approached by the FiLM CNN only around the 35th available day, and keeps a clear margin over the bias-corrected competitors. This shows that the learned, state-dependent bias term goes beyond a climatological correction, and confirms the effectiveness of the framework in exploiting real, extremely sparse observations.

\subsection{Computational Cost}
\label{subsec:comp_cost}

Beyond accuracy, operational deployment requires a limited computational cost, particularly for daily, low-latency updates. The proposed framework is trained on a single NVIDIA RTX 4090 GPU in approximately $1$ hour ($500$ epochs), and the FiLM CNN is trained under the same protocol and hardware, with a cost of the same order. The 3D-Var covariances are instead tuned once, offline, via Bayesian Optimization (Section \ref{experimental_results}), and the classical GP hyperparameters are obtained by maximizing the marginal log-likelihood, both at negligible cost. Since these procedures are not directly comparable to gradient-based training, we focus on inference, which is the recurring cost in operational settings. Inference is evaluated on a resource-constrained CPU (Apple M1 Pro, 16\,GB RAM), without the GPU required for training.

\begin{table}
\centering
\caption{Per-day inference time of the four assimilation approaches as a function of the density of synthetic observations.}
\label{tab:inf_cost}
\begin{tabular}{lcccc}
\hline
\textbf{Obs. Density} & \textbf{3D-Var [s]} & \textbf{GP [s]} & \textbf{FiLM CNN [s]} & \textbf{Proposed [s]} \\
\hline
$1.0\%$ & 17.9385 & 0.0595 & 0.0743 & 0.1617 \\
$1.4\%$ & 16.1644 & 0.0614 & 0.0764 & 0.2776 \\
$1.9\%$ & 17.7259 & 0.0628 & 0.0850 & 0.3038 \\
$2.3\%$ & 16.9607 & 0.0763 & 0.0858 & 0.2836 \\
$2.8\%$ & 18.3246 & 0.0697 & 0.0810 & 0.3082 \\
$3.2\%$ & 16.7395 & 0.0734 & 0.0820 & 0.3156 \\
$3.7\%$ & 18.8173 & 0.0824 & 0.0858 & 0.3413 \\
$4.1\%$ & 16.2498 & 0.0773 & 0.0760 & 0.3209 \\
$4.6\%$ & 16.2388 & 0.0871 & 0.1088 & 0.3113 \\
$5.0\%$ & 16.5836 & 0.0766 & 0.0748 & 0.3155 \\
\hline
\end{tabular}
\end{table}

Inference time depends on the number of observations $m$, which determines the size of the linear system solved at each assimilation step. Table \ref{tab:inf_cost} reports the per-day inference time for observation densities from $1\%$ to $5\%$. 3D-Var is consistently the slowest method, requiring about $16$--$19$\,s per day: although its analysis is also computed in closed form, the minimizer of the cost function \eqref{eq:3dvar_cost} is obtained in state space, through a linear system whose size equals the number of ocean grid points ($n = 9{,}709$), so that its cost is dominated by the full domain rather than by the observations. The proposed method instead solves an $m \times m$ system in observation space, whose cost scales with $m$ rather than with the full domain. The classical GP, operating directly on the raw input features, and the FiLM CNN, whose cost is dominated by a single forward pass, are the fastest methods, with comparable times. The proposed framework adds a modest, roughly constant overhead due to the GP conditioning stage. Although its inference time is several times higher than those of the GP and the FiLM CNN, it remains below one second, an overhead largely outweighed by the accuracy gains discussed in Section \ref{numerical_results}.

\section{Learned Covariance Matrix Analysis}

The kernel parameters, learned jointly with the network weights, define a similarity measure between grid points in the latent space of the neural network: the resulting covariance $\mathbf{K}_{\boldsymbol{\gamma},\mathbf{Z},\mathbf{Z}}$ thus encodes learned correlations among the background errors at different locations. At inference, evaluated on the embeddings $\mathbf{Z}^*$ of the current background state, the covariance $\mathbf{K}_{\boldsymbol{\gamma},\mathbf{Z}^*,\mathbf{Z}^*}$ determines how the information provided by the observations is propagated across the domain. Because similarity is measured in the learned feature space rather than in physical space, two distant grid points can be strongly correlated whenever their embeddings are similar, allowing the model to capture relationships that a stationary spatial kernel cannot represent.

\begin{figure}
    \centering
    \includegraphics[width=\linewidth]{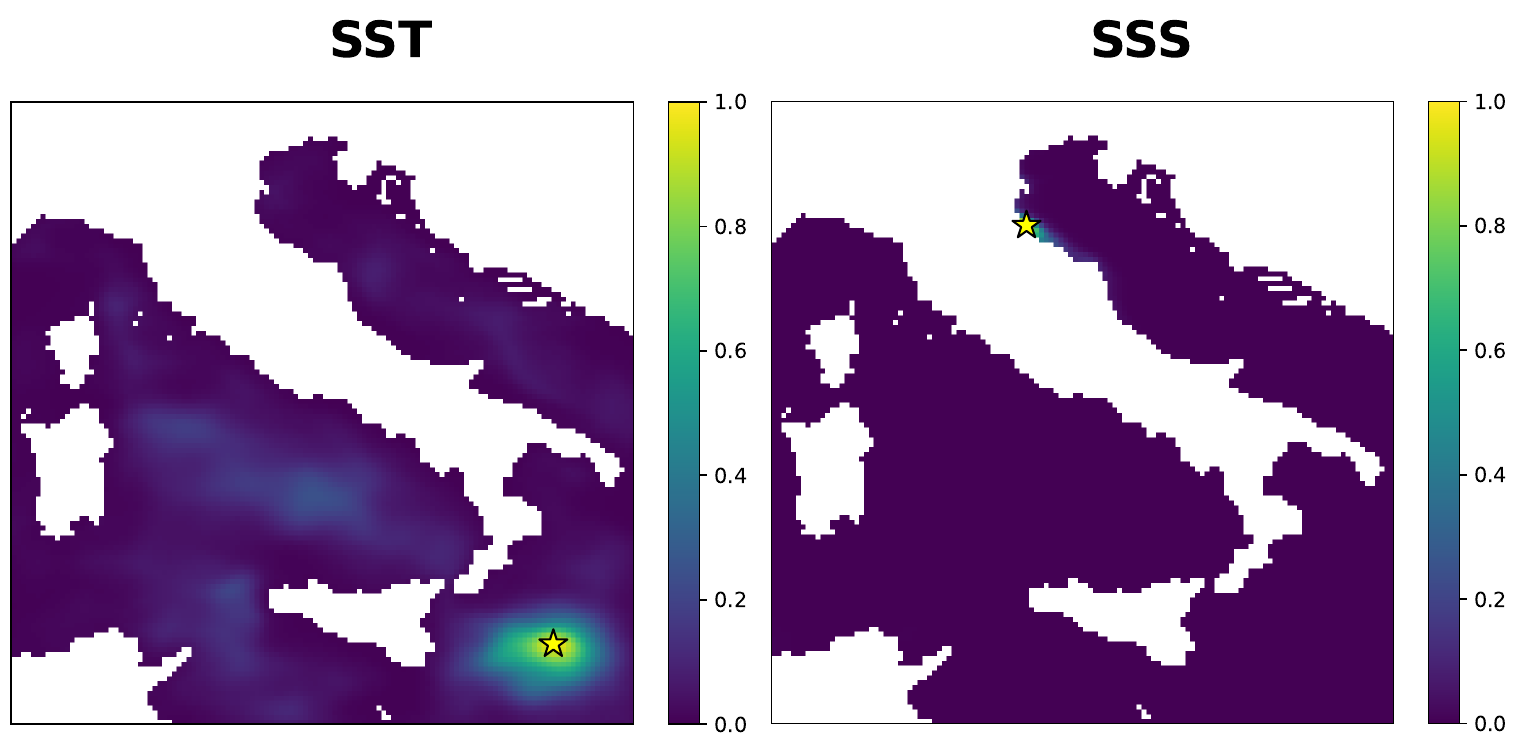}
   \caption{Learned SST and SSS correlations between the location marked by the star and every other point.}
    \label{fig:cov_map}
\end{figure}

To visualize these relationships, Figure~\ref{fig:cov_map} maps a single row of the correlation matrix obtained by normalizing $\mathbf{K}_{\boldsymbol{\gamma},\mathbf{Z}^*,\mathbf{Z}^*}$ back onto the spatial grid. For SST, the open-ocean reference point shows a strong local correlation, surrounded by localized patterns of weaker correlation that extend beyond its immediate neighborhood. For SSS, the model autonomously captures a well-known physical phenomenon: the reference point, located near the Po River outflow, is weakly correlated with the rest of the basin and strongly correlated only within the river plume, where freshwater discharge alters salinity. In both cases, the patterns around the reference point are not circular but take complex, unconstrained shapes learned from the data.

\subsection{Hybrid Data Assimilation}

\begin{figure}
    \centering
    \includegraphics[width=\linewidth]{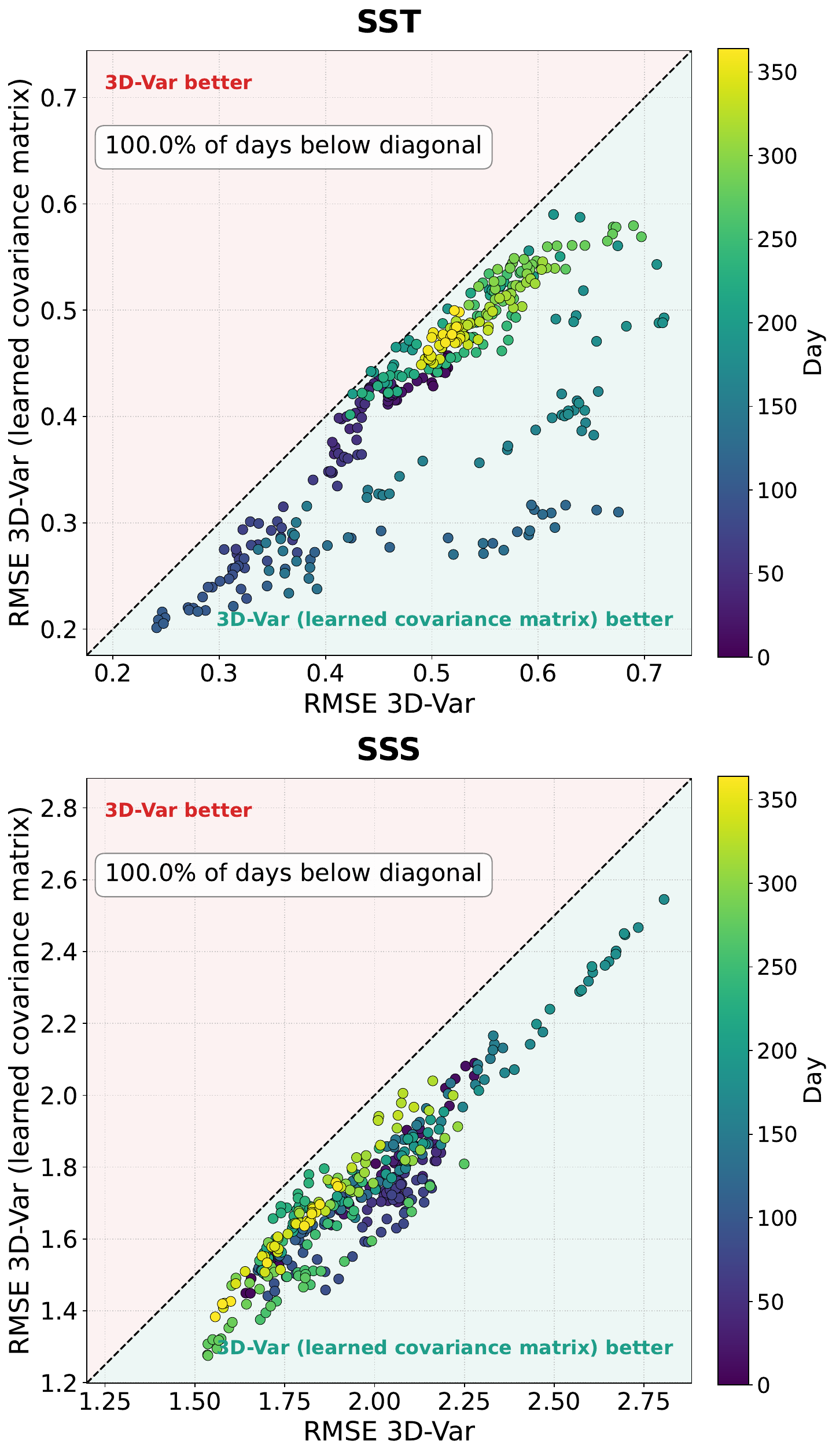}
    \caption{Paired comparison of the daily RMSE obtained by 3D-Var with the classical Gaussian background error covariance (x-axis) and with the covariance learned by the proposed method (y-axis), over the test year for SST and SSS.}
    \label{fig:3dvar_scatter}
\end{figure}

We replace the Gaussian background error covariance $\mathbf{B}$ of the 3D-Var benchmark with the learned covariance $\mathbf{K}_{\boldsymbol{\gamma},\mathbf{Z}^*,\mathbf{Z}^*}$, keeping all other components unchanged (background state, observations and observation error covariance $\mathbf{R}$), so that any difference in performance is due solely to $\mathbf{B}$. Note that, for $\mathbf{R} = \sigma^2\mathbf{I}_m$, the analysis increment of this configuration coincides with the correction of the proposed method with zero prior mean, so that this experiment also isolates the contribution of the learned covariance from that of the learned bias term. Figure~\ref{fig:3dvar_scatter} compares the daily RMSE of the two configurations over the test year for SST and SSS, with $1\%$ of synthetic observations. For both variables, the learned covariance reduces the error on every day of the test year.

For SST, the largest gains occur approximately between days 100 and 200, when the error of the Gaussian configuration is highest and, on several days, is almost halved by the learned covariance. In the first and last months of the year, instead, the points lie closer to the diagonal, indicating smaller but consistent improvements. The full proposed method still performs better, since it also benefits from the learned prior mean, which captures the systematic bias, but the learned $\mathbf{B}$ alone already improves the structure of the 3D-Var corrections.

For SSS, the points lie along a narrow band parallel to the diagonal, indicating a nearly constant improvement regardless of error level and season.

Overall, the learned covariance is not only an interpretable product of the framework but also a better background error covariance in its own right, which can be plugged directly into an existing variational system without modifying its structure. The proposed framework therefore serves a dual purpose: an effective covariance generator for DA and an interpretability tool for investigating the spatial structure of the background error. 
\section{Conclusions}
In this work, we presented a hybrid data assimilation framework that combines neural networks and Gaussian Processes, trained end-to-end by maximizing the marginal log-likelihood. A CNN feature extractor generates embeddings that parametrize a learnable GP kernel. This overcomes two main limitations of classical variational approaches such as 3D-Var: a hand-crafted, static background error covariance and the lack of an explicit bias correction. Unlike purely data-driven solutions, the framework remains interpretable and provides uncertainty quantification by design, since both the learned covariance and the predictive variance are direct outputs of the GP rather than post-hoc approximations. A spatial patching strategy addresses the $\mathcal{O}(n^3)$ cost of exact GP training, allowing the framework to scale to large, high-resolution domains with a controllable trade-off between cost and accuracy.

Experiments on SST and SSS, with both synthetic observations at controlled densities and real, irregularly distributed buoy measurements, showed that the proposed approach consistently outperforms 3D-Var, a classical GP and a FiLM-conditioned CNN. Even with real buoys only ($0.03\%$ and $0.06\%$ observation density), it achieves effective assimilation and implicit bias correction, also outperforming competitors equipped with a climatological bias correction. The learned covariances capture state-dependent spatial structures, such as the Po River plume, that a stationary kernel cannot represent. Moreover, when plugged into a classical 3D-Var scheme in place of the hand-crafted Gaussian covariance, without further tuning, the learned covariance reduces the error on all test days for both SST and SSS in the synthetic setting. Operational systems can thus benefit from data-driven covariance modeling while preserving their variational formulation. This also opens the way to settings where closed-form GP inference is unavailable, such as non-linear observation operators, additional physical constraints or 4D-Var.

Overall, combining deep learning with Gaussian Processes offers an effective, interpretable and scalable alternative to traditional DA, as well as a practical means of enhancing it, particularly in operational settings with extreme observational sparsity.

\bibliographystyle{IEEEtran}
\bibliography{bibliography}

\end{document}